\PassOptionsToPackage{table}{xcolor}
\documentclass[runningheads]{llncs}

\usepackage[mobile]{eccv}

\usepackage{colortbl}

\usepackage{eccvabbrv}

\usepackage{graphicx}
\usepackage{booktabs}
\makeatletter
\usepackage{subcaption}
\newcommand{\makecell}[2][c]{%
  \if#1l\begin{tabular}[t]{@{}l@{}}#2\end{tabular}%
  \else\if#1r\begin{tabular}[t]{@{}r@{}}#2\end{tabular}%
  \else\begin{tabular}[t]{@{}c@{}}#2\end{tabular}\fi\fi}
\makeatother
\usepackage{multirow}
\usepackage{wrapfig}

\usepackage[accsupp]{axessibility}  
\newcommand{\cmark}{\checkmark}
\newcommand{\xmark}{$\times$}

\usepackage{hyperref}

\usepackage{orcidlink}
\definecolor{deepgreen}{RGB}{0,100,0}

\begin{document}

\title{Z-FLoc: Zero-Shot Floorplan Localization \\ via Geometric Primitives} 

\titlerunning{Z-FLoc: Zero-Shot Floorplan Localization via Geometric Primitives}

\author{Ayumi Umemura\inst{1,2}, Toshinori Kuwahara\inst{2}, Marc Pollefeys\inst{1}, Daniel Barath\inst{1}}
\authorrunning{A. Umemura et al.}
\definecolor{softgreen}{RGB}{76,175,80}
\institute{ETH Zurich\and
Tohoku University}
\definecolor{gray}{RGB}{100,100,100}

\maketitle

\begin{abstract}
  Visual localization -- estimating a camera pose within a pre-existing map -- is a fundamental problem in computer vision.
  Floorplans are an attractive map representation: they are readily available for most buildings, compact, and inherently invariant to visual appearance changes.
  However, bridging the severe domain gap between camera observations and floorplan geometry remains challenging.
  Existing methods address this gap through data-driven learning, yet they require large-scale training data and environment-specific retraining, limiting their practical deployment.
  We propose a zero-shot floorplan localization method that generalizes to novel environments \textit{without} any retraining.
  Our key insight is that dominant geometric primitives -- lines and circles -- are ubiquitous in human-made environments and provide appearance-invariant structural constraints.
  We extract these primitives from a bird's-eye-view (BEV) projection of monocular 3D reconstructions and match them to the floorplan via dedicated minimal solvers within a robust estimation framework.
  Experiments on both simulated and real-world datasets show that our approach outperforms state-of-the-art learning-based methods on unseen environments, while using a single fixed set of hyperparameters across all experiments.
  The source code will be made publicly available.
  
\end{abstract}

    \vspace{-3mm}
\section{Introduction}
\label{sec:intro}
    \vspace{-1mm}

Camera localization is a core problem in computer vision that underpins applications ranging from augmented reality on mobile devices to autonomous robotic navigation.
Conventional approaches localize by matching against a pre-acquired image database~\cite{city_scale_location_recognition, anyloc, revisit_anything} or within a reconstructed 3D model~\cite{hloc, ace, lindenberger2023lightglue}.
Despite their accuracy, such representations demand substantial effort for data collection, storage, and long-term maintenance -- making them impractical for many real-world deployments.

Floorplans offer a compelling alternative.
Most indoor environments already provide them as part of their building infrastructure, eliminating additional data acquisition costs.
A floorplan compactly encodes the structural layout of a space and is inherently invariant to appearance changes caused by furniture, lighting, or decoration.
In this work, we localize a monocular camera with respect to a given 2D floorplan, enabling practical indoor localization without LiDAR or environment-specific 3D maps.

The central challenge lies in establishing correspondences between camera observations and floorplan geometry.
Two factors make this particularly difficult: (i) a severe \emph{modality gap} -- images versus schematic top-down drawings -- and (ii) fundamentally different coordinate frames -- egocentric versus allocentric.
Prior work has predominantly addressed this gap through learning~\cite{pfnet, laser, f3loc, c3po, unloc}, achieving strong accuracy but requiring extensive training data and environment-specific retraining, which limits generalization.
Training-free alternatives~\cite{map_alignment, graph_floorloc} sidestep the data requirement but instead rely on high-precision, high-coverage point clouds, which are equally restrictive in practice.

We propose a zero-shot floorplan localization framework that overcomes both limitations.
Our key insight is that structural primitives -- lines and circles -- are ubiquitous in human-made environments and induce stable orientation and distance constraints that persist even under noisy, partial observations.
Building on this observation, we reconstruct a bird’s-eye-view (BEV) map from monocular images, extract 2D line and circle primitives, and match them to the floorplan using a suite of dedicated minimal solvers embedded in a robust hypothesis-and-verify estimation framework.
A complementary scoring function combining alignment consistency with free-space violation rejection selects the best hypothesis, which is then refined via nonlinear optimization.
Experiments on both simulated and real-world datasets show that our method generalizes reliably across diverse, unseen environments -- outperforming state-of-the-art learning-based methods that were specifically trained on the target domain -- all with a single, fixed set of hyperparameters.

    \vspace{-2mm}
\section{Related Work}
    \vspace{-1mm}

\textbf{Visual localization} estimates camera poses with respect to a geo-referenced representation.
Conventional approaches~\cite{hloc, learning_to_semidense, feature_to_point_matching, pixloc, wang2023dgc} build Structure-from-Motion (SfM) maps and establish 3D--2D correspondences for Perspective-\textit{n}-Point (PnP) pose estimation~\cite{epnp}.
Scene coordinate regression~\cite{apr_1} and direct pose regression~\cite{yin2025towards} methods bypass explicit correspondences by predicting poses with deep networks.
While highly accurate, both paradigms require pre-built 3D maps and large-scale data collection.
An alternative line of work localizes against overhead representations such as satellite imagery or floorplans~\cite{lalaloc, f3loc, geometric_loc, palms+}, estimating planar $\mathrm{SE}(2)$ poses without a full 3D map.

\vspace{1mm}
\noindent
\textbf{Floorplan localization} bridges the gap between perspective camera views and sparse, top-down structural drawings.
\emph{Learning-based methods.}
LaLaLoc~\cite{lalaloc, lalaloc++} embeds map and image features into a shared latent space for feature-level comparison.
Several methods~\cite{pfnet, f3loc, semantic_f3loc} combine learned observation models with sequential filtering to accumulate evidence over time.
C3Po~\cite{c3po} leverages DUSt3R~\cite{dust3r} to learn point-level correspondences between images and floorplans, casting localization as a geometric registration problem.
FloorplanNet~\cite{floorplan_net} matches LiDAR-derived occupancy maps to floorplans with graph neural networks.
A complementary line leverages semantic scene understanding: SeDAR~\cite{sedar} replaces depth with CNN-predicted semantic labels for Monte Carlo localization against floorplans, SPVLoc~\cite{spvloc} matches perspective image embeddings to semantic panoramas rendered from untextured 3D models, and OrienterNet~\cite{orienternet} aligns neural bird's-eye-view predictions with OpenStreetMap for outdoor localization.
Despite their accuracy, all of these methods require domain-specific training data and must be retrained for each new environment.

\vspace{1mm}
\emph{Training-free methods.}
Ewe~\etal~\cite{graph_floorloc} estimate similarity transformations via SIFT~\cite{sift} matching on LiDAR-derived maps.
Other works~\cite{area_graph, map_alignment} detect room shapes in occupancy grids and match handcrafted descriptors.
These approaches avoid training but depend on high-quality, high-coverage point clouds to reconstruct complete room geometries.
Kim~\etal~\cite{geometric_loc} propose a fully geometric method based on 2D--3D line correspondences, using a distance field over image line segments for pose search followed by primitive-based refinement.
While demonstrating strong generalization, their approach requires panoramic images and accurate height information.
PALMS+~\cite{palms+} reconstructs point clouds with an off-the-shelf depth foundation model and localizes via BEV-to-floorplan correlation, but requires a full 360$^\circ$ capture from a fixed viewpoint.

In summary, existing floorplan localization methods face one of two fundamental limitations:
(1)~data-driven approaches require retraining for each new environment, and
(2)~model-free approaches impose strict requirements on sensor data, particularly observation completeness and measurement accuracy.
Our method addresses both: it generalizes across environments without retraining and maintains high accuracy under incomplete, noisy observations -- all with a single, fixed set of hyperparameters.

    \vspace{-2mm}
\section{Zero-Shot Floorplan Localization}
    \vspace{-1mm}

\textbf{Problem formulation.}
Given a sequence of RGB images $\{I_i\}_{i=1}^N$ with known relative $\mathrm{SE}(3)$ poses and intrinsics, we seek the $\mathrm{SE}(2)$ pose of a monocular camera within a given 2D floorplan.
We cast this as a cross-modal matching problem: a BEV map reconstructed from the images must be aligned to the floorplan via a similarity transformation parameterized by scale $s$, rotation $\mathbf{r} \in \mathrm{SO}(2)$, and translation $\mathbf{t} \in \mathbb{R}^2$.
The floorplan provides only geometric layout information (\eg, walls and structural boundaries) without height or semantic annotations.

\begin{figure}[t]
    \centering
    \includegraphics[width=\linewidth]{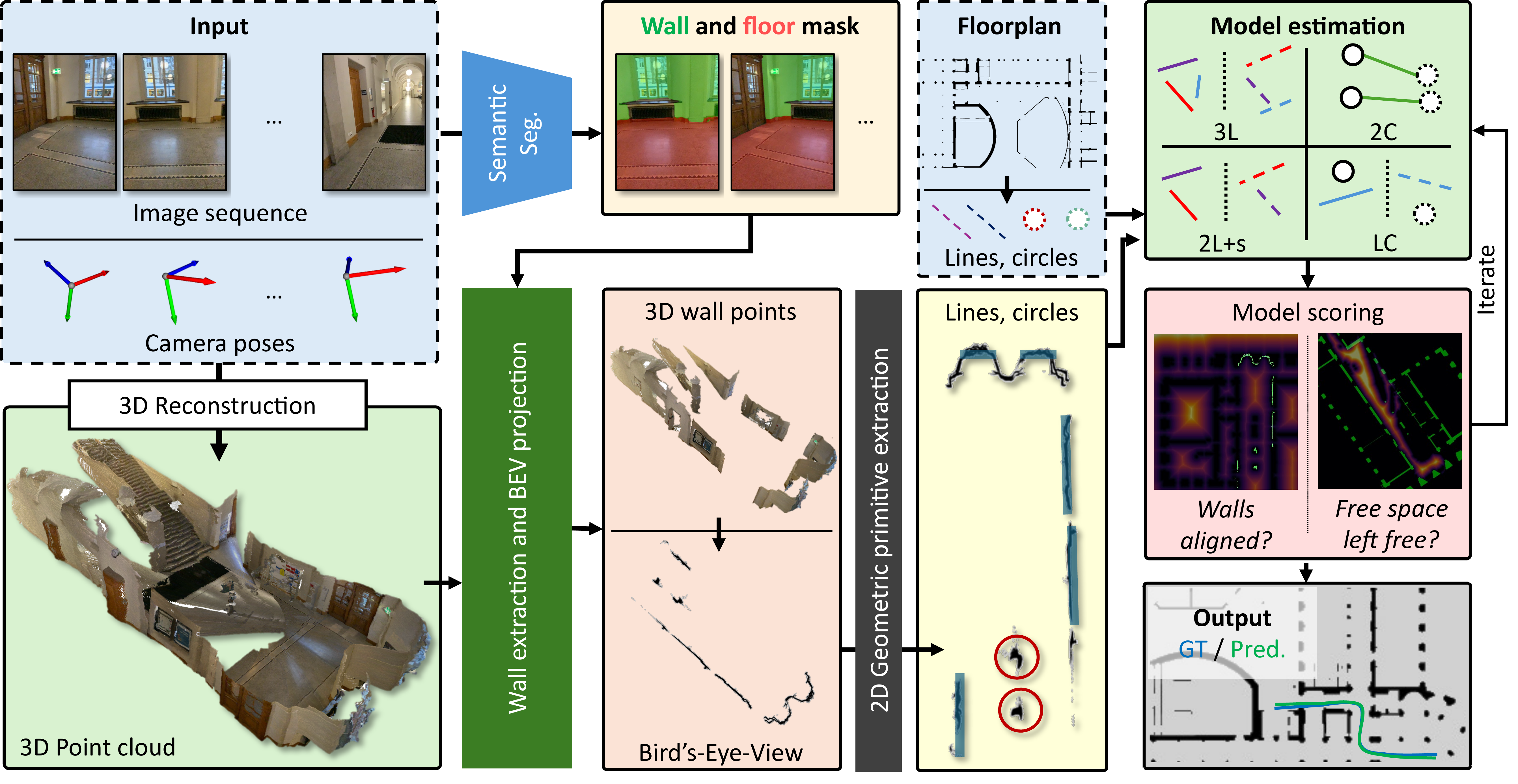}
    \caption{\textbf{Pipeline overview.} Given a sequence of images, we reconstruct wall-only 3D points and project them along gravity to obtain a bird's-eye-view (BEV) map. Line segments and circular primitives are extracted from the BEV and matched to floorplan primitives via dedicated minimal solvers within a hybrid RANSAC loop. Each hypothesis is scored by two complementary cost functions -- a consistency score rewarding BEV-to-floorplan alignment and a violation score penalizing overlap with observed free space -- and the top-ranked hypothesis is refined via nonlinear optimization.}
    \vspace{-3mm}
    \label{fig:overview}
\end{figure}

\vspace{1mm} \noindent
\textbf{Pipeline overview.}
We first reconstruct wall-only 3D points and project them along gravity to obtain a BEV map, from which line segments and circular primitives are extracted.
Candidate similarity transformations are generated from randomly sampled minimal sets of primitives -- line--line, circle--circle, and line--circle pairs -- each yielding an independent pose hypothesis.
Hypotheses are scored by a complementary pair of cost functions that reward alignment consistency and penalize free-space violations.
The top-ranked hypothesis is refined via nonlinear optimization.
An overview is shown in Fig.~\ref{fig:overview}.

    \vspace{-1mm}
\subsection{Top-View Geometric Primitive Extraction}
    \vspace{-1mm}
    
We reconstruct 3D scene points from the input image sequence using a visual foundation model (\eg, MapAnything~\cite{mapanything}), conditioned on the known relative camera poses and intrinsics for geometric consistency.
These poses can be estimated via any off-the-shelf visual or visual-inertial SLAM system, depending on our capturing device (\eg, smartphones are typically equipped with high-quality IMUs). 
To obtain a BEV representation, we segment wall and floor regions with Mask2Former~\cite{mask2former}, suppressing clutter from occluding objects such as furniture.
The gravity direction is estimated by fitting a plane via PCA~\cite{pca_review} to the floor point cloud; its downward normal defines the gravity vector.

\begin{figure}[t]
    \centering
    \begin{subfigure}[t]{0.135\linewidth}
        \centering
        \includegraphics[width=\linewidth]{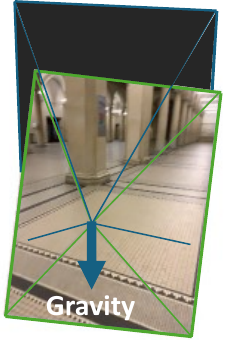}
        \caption{}
    \end{subfigure}
    \hfill
    \begin{subfigure}[t]{0.55\linewidth}
        \centering
        \includegraphics[width=\linewidth]{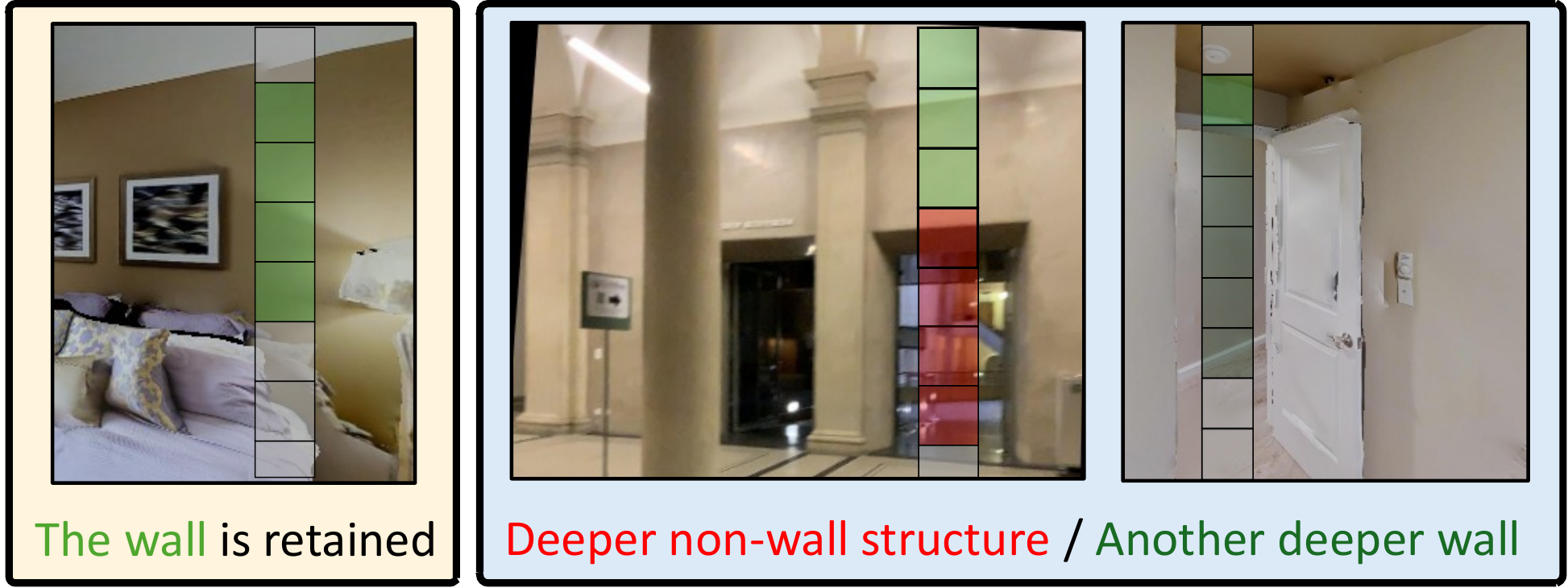}
        \caption{}
    \end{subfigure}
    \hfill
    \begin{subfigure}[t]{0.255\linewidth}
        \centering
        \includegraphics[width=\linewidth]{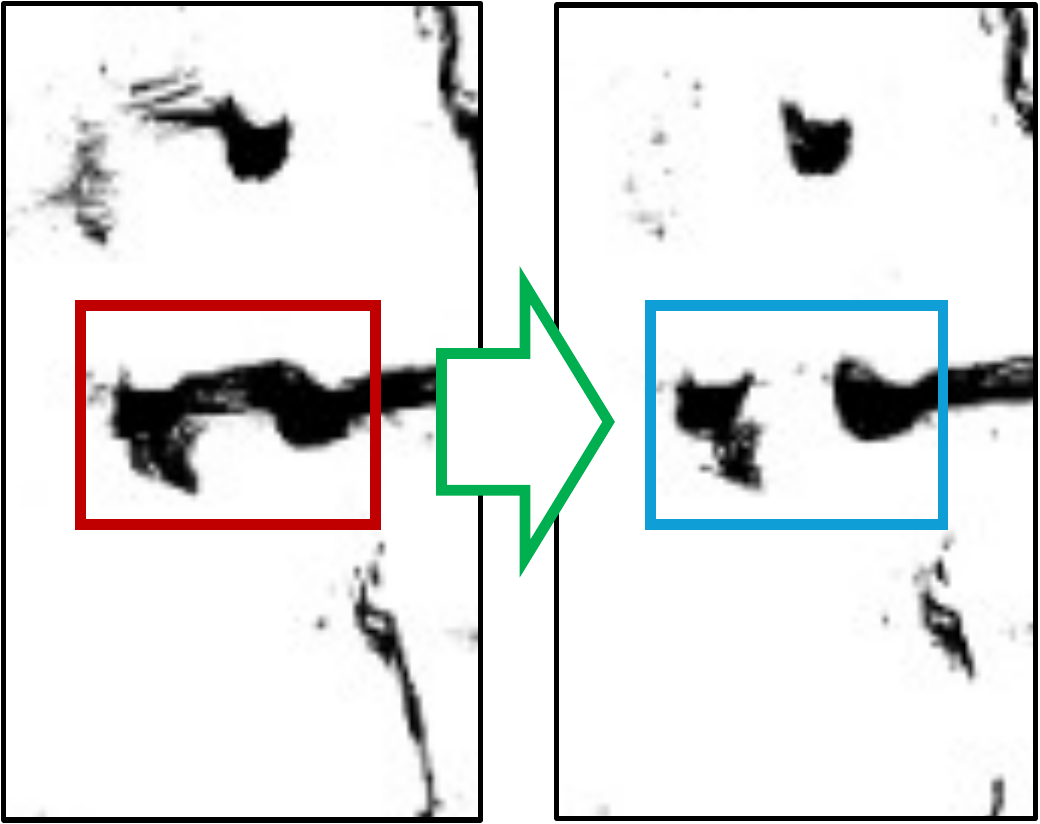}
        \caption{}
    \end{subfigure}
    \vspace{-2mm}
    \caption{\textbf{Wall mask filtering.} (a)~The input image is warped into a gravity-aligned frame so that vertical walls produce consistent depth values per column. (b)~Column-wise filtering: if any non-wall pixel in a column has depth exceeding the estimated \textcolor{softgreen}{wall depth}, the column is discarded (\textcolor{red}{red} pixels); when multiple wall structures are present, the \textcolor{deepgreen}{deeper wall surface} is retained. (c)~Filtering suppresses ghost artifacts from segmentation errors and preserves traversable openings such as doors and arches.}
    \label{fig:wall_mask}
    \vspace{-5mm}
\end{figure}

\vspace{1mm}
\noindent
\textbf{Wall mask refinement.}
Segmentation errors may label non-structural regions as walls, producing ghost artifacts in the BEV.
We suppress these by exploiting two geometric properties of gravity-aligned images:
(1)~vertical walls induce a dominant depth value per image column, and
(2)~non-wall regions within the same column generally lie \emph{closer} to the camera than the wall surface.
We first warp the image into a gravity-aligned frame. Given intrinsics $\mathbf{K}$ and rotation $\mathbf{R} \in \mathrm{SO}(3)$, the warping homography is
\begin{equation}
\hat{\mathbf{p}} = \mathbf{K}\mathbf{R}\mathbf{K}^{-1}\mathbf{p},
\label{eq:gravity_align}
\end{equation}
where $\mathbf{p}$ and $\hat{\mathbf{p}}$ are homogeneous pixel coordinates.
We apply Eq.~\ref{eq:gravity_align} to the wall-mask pixels and their corresponding depth maps.
The gravity-aligned depth is
\begin{equation}
\hat{d} =
\left[
\mathbf{R}^{T}
\left(
\mathbf{K}^{-1}\mathbf{p}\, d
\right)
\right]_z ,
\end{equation}
where $d$ is the original depth and $[\cdot]_z$ denotes the $z$-component of the resulting 3D vector.
After gravity alignment, vertical walls exhibit consistent depth values within each image column (Fig.~\ref{fig:wall_mask}a, b).
The representative wall depth for column $c$ is
\begin{equation}
\tilde{d}_c = \mathrm{median}\left\{ \hat{d}(\mathbf{p}) \;\middle|\; \mathbf{p} \in \mathcal{M}_c \right\},
\end{equation}
where $\mathcal{M}_c$ denotes the set of wall-mask pixels in column $c$.

If any non-wall pixel in a column has depth exceeding the wall depth, the entire column is likely an erroneous detection.
The refined wall mask for column $c$ is as follows:
\begin{equation}
\mathcal{M}_c^{\mathrm{ref}} =
\begin{cases}
\mathcal{M}_c
& \text{if }
\max_{\mathbf{p} \notin \mathcal{M}_c}
\hat{d}(\mathbf{p})
\le
\tilde{d}_c + \tau_d, \\
\emptyset
& \text{otherwise},
\end{cases}
\end{equation}
where $\tau_d$ is a distance threshold.
Moreover, if multiple wall instances appear within a column, we keep only the farthest wall segment. 
Let $\{\mathcal{S}_{c,k}\}$ denote depth-separated wall segments in column $c$. 
The refined wall mask is defined as
\begin{equation}
k^\star =
\arg\max_k
\ \mathrm{median}_{\mathbf{p}\in\mathcal{S}_{c,k}} \hat{d}(\mathbf{p}),
\qquad
\mathcal{M}_c^{\mathrm{ref}} = \mathcal{S}_{c,k^\star}.
\end{equation}
This retains the farthest wall structure while discarding shallower detections.

Points in $\mathcal{M}_c^{\mathrm{ref}}$ are projected onto the gravity-orthogonal plane to produce the BEV.
While this filtering can occasionally be overly conservative, we deliberately favor precision: sequential observations recover missed wall regions from other viewpoints, whereas unfiltered noise produces persistent ghost artifacts.

\vspace{1mm}
\noindent
\textbf{Line extraction.}
Line segments are extracted from the BEV wall pixels via seed-based region growing.
Unassigned pixels are randomly sampled as anchors; for each anchor, PCA on its local neighborhood yields a dominant direction that initializes a line model.
The region is iteratively expanded by including neighboring pixels whose orthogonal distance to the current line falls below a threshold, with the line parameters re-estimated by least squares after each step.
Growth terminates when no further pixels qualify, and the procedure repeats until all wall pixels are assigned.

\vspace{1mm}
\noindent
\textbf{Circle extraction.}
Indoor environments often contain closed structures such as cylindrical pillars that provide useful matching constraints.
We cluster the BEV wall pixels with HDBSCAN~\cite{hdbscan}; spatially isolated structures naturally form separate clusters.
For each cluster, a circle (center and radius) is fitted via RANSAC with minimal three-point samples.
A cluster is accepted as a valid circle if it satisfies two criteria:
(i)~\emph{angular coverage} -- the inlier points must subtend a sufficiently large arc around the center, measured as $2\pi$ minus the largest angular gap between consecutive inliers; and
(ii)~\emph{angular continuity} -- consecutive angular differences must remain below a gap threshold, and the contiguous span must exceed a minimum extent.

\subsection{Model Estimation}
\label{sec:solvers}
We derive a family of minimal solvers for the 2D similarity transform between the BEV and the floorplan, and present a hypothesis pruning strategy for efficient estimation.

\vspace{1mm} \noindent
\textbf{Three-line (3L) solver.}
A 2D line $l$ is represented in normal form as
\begin{equation}
\mathbf{n}^\top \mathbf{x} + d = 0,
\quad \|\mathbf{n}\|_2 = 1,
\end{equation}
where $\mathbf{n}$ is the unit normal, $d$ the signed offset, and $\mathbf{x}$ a point on the line.
The 2D similarity transform from BEV to floorplan is
\begin{equation}
\mathbf{x}' = s \mathbf{r}\mathbf{x} + \mathbf{t},
\end{equation}
with scale $s$, rotation $\mathbf{r}\!\in\!\mathrm{SO}(2)$, and translation $\mathbf{t}\!\in\!\mathbb{R}^2$.
Under this transformation, the line parameters become
\begin{equation}
\label{eq:trans_line_param}
\mathbf{n}' = \mathbf{r}\mathbf{n},
\qquad
d' = s d - \mathbf{n}'^\top \mathbf{t}.
\end{equation}
Given three line correspondences
$\{(l_i, l'_i)\}_{i=1}^3$,
the rotation is determined from a single pair by aligning normals,
$\mathbf{n}'_i = \mathbf{r}\mathbf{n}_i$,
up to a $\pi$ ambiguity (since $\mathbf{n}$ and $-\mathbf{n}$ represent the same line); we consider both candidates.
Fixing $\mathbf{r}$, the remaining unknowns $(s, \mathbf{t})$ follow from the $3\!\times\!3$ linear system
\begin{equation}
\begin{bmatrix}
d_1 & \mathbf{n}'_1{}^\top \\
d_2 & \mathbf{n}'_2{}^\top \\
d_3 & \mathbf{n}'_3{}^\top
\end{bmatrix}
\begin{bmatrix}
s \\ -\mathbf{t}
\end{bmatrix}
=
\begin{bmatrix}
d'_1 \\ d'_2 \\ d'_3
\end{bmatrix}.
\end{equation}
The system is solvable whenever the three lines are not mutually parallel.

\vspace{1mm} \noindent
\textbf{Two-line (2L) solver.}
When an approximately metric scale estimate $\tilde{s}$ is available from the reconstruction backbone (\eg, MapAnything), two line correspondences suffice.
To account for scale uncertainty, we perturb $\tilde{s}$ with zero-mean generalized Gaussian noise
$\epsilon \sim \mathrm{GGD}(0,\sigma,4)$,
where $\sigma$ is set so that
$\mathbb{P}(|\epsilon|\le 0.1\tilde{s})\approx0.8$,
yielding $s=\tilde{s}+\epsilon$.
The rotation is obtained from a single normal alignment as above, and the translation from
\begin{equation}
\mathbf{n'_i}{}^\top\mathbf{t} = sd_i-d_i', \quad i\in\{1,2\}.
\end{equation}
The system is solvable when the two lines are not parallel.

\vspace{1mm} \noindent
\textbf{Two-circle (2C) solver.}
Given two pairs of corresponding circle centers
$(\mathbf{c}_1, \mathbf{c}'_1)$ and $(\mathbf{c}_2, \mathbf{c}'_2)$,
the similarity transform is uniquely determined.
The scale is the ratio of inter-center distances as follows:
\begin{equation}
s = \frac{\|\mathbf{c}'_2 - \mathbf{c}'_1\|_2}{\|\mathbf{c}_2 - \mathbf{c}_1\|_2}.
\end{equation}
Letting
$\mathbf{v} = \mathbf{c}_2 - \mathbf{c}_1$ and
$\mathbf{v}' = \mathbf{c}'_2 - \mathbf{c}'_1$,
the rotation angle and translation are
\begin{equation}
\theta = \mathrm{atan2}(v'_y, v'_x) - \mathrm{atan2}(v_y, v_x),
\quad
\mathbf{r} =
\begin{bmatrix}
\cos\theta & -\sin\theta \\
\sin\theta & \cos\theta
\end{bmatrix}, \quad
\mathbf{t} = \mathbf{c}'_1 - s \mathbf{r} \mathbf{c}_1.
\end{equation}

\vspace{1mm} \noindent
\textbf{Line-circle (LC) solver.}
Given a line correspondence $(l, l')$
and a circle-center correspondence $(\mathbf{c}, \mathbf{c}')$,
the rotation follows from normal alignment ($\mathbf{n}'=\mathbf{r}\mathbf{n}$, up to $\pi$).
Substituting into the line-offset constraint yields the scale,
\begin{equation}
s =
\frac{d' + \mathbf{n}'^\top \mathbf{c}'}
     {d + \mathbf{n}'^\top \mathbf{r}\mathbf{c}},
\end{equation}
and the translation follows as
$\mathbf{t} = \mathbf{c}' - s\mathbf{r}\mathbf{c}$.

\vspace{1mm}
\noindent
\textbf{Hybrid solver selection.}
The four solvers above address complementary primitive configurations.
We combine them in a hybrid RANSAC loop~\cite{hybrid_ransac}: at each iteration, a solver is randomly selected from the enabled pool, a minimal sample of the corresponding primitive type is drawn, and the resulting similarity hypothesis is scored against all primitives jointly.
This strategy avoids committing to a single primitive type and lets the estimation adapt to whichever geometric cues are most informative in a given scene.

\vspace{1mm} \noindent
\textbf{Hypothesis pruning for the 3L solver.}
Without learned descriptors, correspondences must be enumerated exhaustively, which becomes prohibitive as the number of floorplan lines grows.
We apply three pruning strategies.
\emph{(i)~Line deduplication}: near-duplicate lines with similar orientation and offset are suppressed in both the floorplan and BEV, retaining longer segments.
\emph{(ii)~Spatial proximity}: since the BEV represents a local region, we restrict sampled floorplan triplets to spatially neighboring segments (inter-line distance is precomputed), reducing combinatorial growth from cubic to quasi-linear in scene size. Degenerate configurations (\eg, all parallel lines) are rejected.
\emph{(iii)~Normal consistency}: after estimating a rotation $\mathbf{r}$ from one correspondence, the transformed normals of all three floorplan lines must be parallel to their BEV counterparts.
We therefore enforce
\begin{equation}
\forall i \in \{1,2,3\}, \quad 
\left| (\mathbf{r}\mathbf{n}_i)^\top \mathbf{n}'_i \right| > \tau_p,
\end{equation}
where $\tau_p$ is a cosine-similarity threshold.
Inconsistent hypotheses are discarded before solving for scale and translation.

    \vspace{-1mm}
\subsection{Model Scoring}
\label{sec:hypothesis}
    \vspace{-1mm}

A correct transformation should produce geometrically consistent alignment between the BEV wall pixels and the floorplan.
We evaluate each hypothesis, estimated from a minimal sample, with two complementary cost functions.

\vspace{1mm} \noindent
\textbf{Consistency score $C^c$.}
Let $X_w$ denote the BEV wall pixels and $X'_w$ the floorplan wall pixels.
The nearest-neighbor distance is
\begin{equation}
d_{\min}(x_1, X_2)
=
\min_{x_2 \in X_2}
\|x_1 - x_2\|_2.
\label{eq:min_func}
\end{equation}
The consistency score measures the fraction of transformed BEV points that fall within distance $\tau_c$ of the floorplan:
\begin{equation}
\label{eq:consistency}
C^c(s, \mathbf{r}, \mathbf{t})
=
\frac{1}{|X_w|}
\sum_{x_w \in X_w}
\rho^c\!\left(
d_{\min}(\mathcal{T}(x_w; s, \mathbf{r}, \mathbf{t}),\; X_w')
\right),
\end{equation}
where $\mathcal{T} \in \mathrm{Sim}(2)$ maps BEV points to the floorplan frame and
$\rho^c(x)=\mathbf{1}(|x|<\tau_c)$ is an indicator function.

\vspace{1mm} \noindent
\textbf{Violation score $C^v$.}
This term penalizes hypotheses that place floorplan walls inside observed free space.
Let $X_{\mathrm{free}}$ denote the observed free space in the BEV and $\mathcal{T}^{-1}$ the inverse transform.
The violation score is as follows:
\begin{equation}
\label{eq:violation}
\begin{gathered}
\tilde{X}_w' = \mathcal{T}^{-1}(X_w'; s,\mathbf{r},\mathbf{t}), \quad
\tilde{X}'_{\mathrm{free}}=\tilde{X}'_{w} \cap X_{\mathrm{free}},
\\[3pt]
C^v(s, \mathbf{r}, \mathbf{t})
=
\frac{1}{|\tilde{X}'_{\mathrm{free}}|}
\sum_{\tilde{x}_w' \in \tilde{X}'_{\mathrm{free}}}
\rho^v\!\left(
d_{\min}(\tilde{x}_w', X_w)
\right),
\end{gathered}
\end{equation}
where $\rho^v(x)=\mathbf{1}(|x|>\tau_v)$.
The final hypothesis score combines both terms as:
\begin{equation}
\label{eq:final_score}
C(s, \mathbf{r}, \mathbf{t})=C^c(s, \mathbf{r}, \mathbf{t})-C^v(s, \mathbf{r}, \mathbf{t}).
\end{equation}

\vspace{1mm} \noindent
\textbf{Efficient score computation.}
Since Eq.~\ref{eq:min_func} requires repeated nearest-neighbor queries, we precompute two distance fields.
The \emph{wall distance field} on the floorplan is calculated as follows:
\begin{equation}
\label{eq:dist_func}
f^c(x'; X'_w) = d_{\min}(x', X'_w),
\end{equation}
and it stores the closest distance from any query point $x'$ to the floorplan walls $X'_w$.
The \emph{clearance field} is built from the BEV occupancy map obtained by projecting all reconstructed 3D points along gravity.
Within the observed free space $X_{\mathrm{free}}$, it stores the distance to the nearest occupied pixel $X_{\mathrm{occ}}$:
\begin{equation}
f^v(x; X_{\mathrm{occ}})
=d_{\min}(x, X_{\mathrm{occ}}),
\qquad
x \in X_{\mathrm{free}}.
\end{equation}
Scoring follows a coarse-to-fine strategy: hypotheses are first ranked by $C^c$ using sparsely sampled BEV line points for fast preselection, then re-ranked using all wall pixels and the full score $C$ (Eq.~\eqref{eq:final_score}).

\subsection{Model Refinement}
\label{sec:refinement}
The top-ranked hypothesis is refined by minimizing
\begin{equation}
\min_{s,\mathbf{r},\mathbf{t}}
\sum_{x_w\in X_w} \phi\!\left(d_{\min}(\mathcal{T}(x_w; s, \mathbf{r}, \mathbf{t}),\; X_{w}')\right),
\end{equation}
where $\phi(\cdot)$ is a Huber loss.
The optimization uses Levenberg--Marquardt; for efficiency, $d_{\min}$ is evaluated leveraging a lookup table in the precomputed distance field $f^c$ (Eq.~\ref{eq:dist_func}).

    \vspace{-2mm}
\section{Experiments}
    \vspace{-1mm}

\noindent
\textbf{Datasets.}
We evaluate on three benchmarks spanning synthetic and real-world indoor environments.
\emph{Gibson(t)}~\cite{igibson} consists of 118 synthetic trajectories (280--5{,}152 frames each) in environments smaller than 300\,m$^2$, captured with upright cameras and a wide field of view (108$^\circ$). We follow the evaluation protocol of~\cite{f3loc}.
\emph{LaMAR}~\cite{sarlin2022lamar} provides real-world indoor sequences captured in two university buildings -- HGE and CAB -- comprising 16 sessions totaling 5{,}187 images over approximately 22{,}500\,m$^2$.
The sequences feature challenging conditions including severe occlusions and a narrow field of view (48$^\circ$).
For HGE, we evaluate two floorplan settings (full and cropped) that vary the spatial coverage.

\begin{figure}[t]
    \centering
    \includegraphics[width=\linewidth]{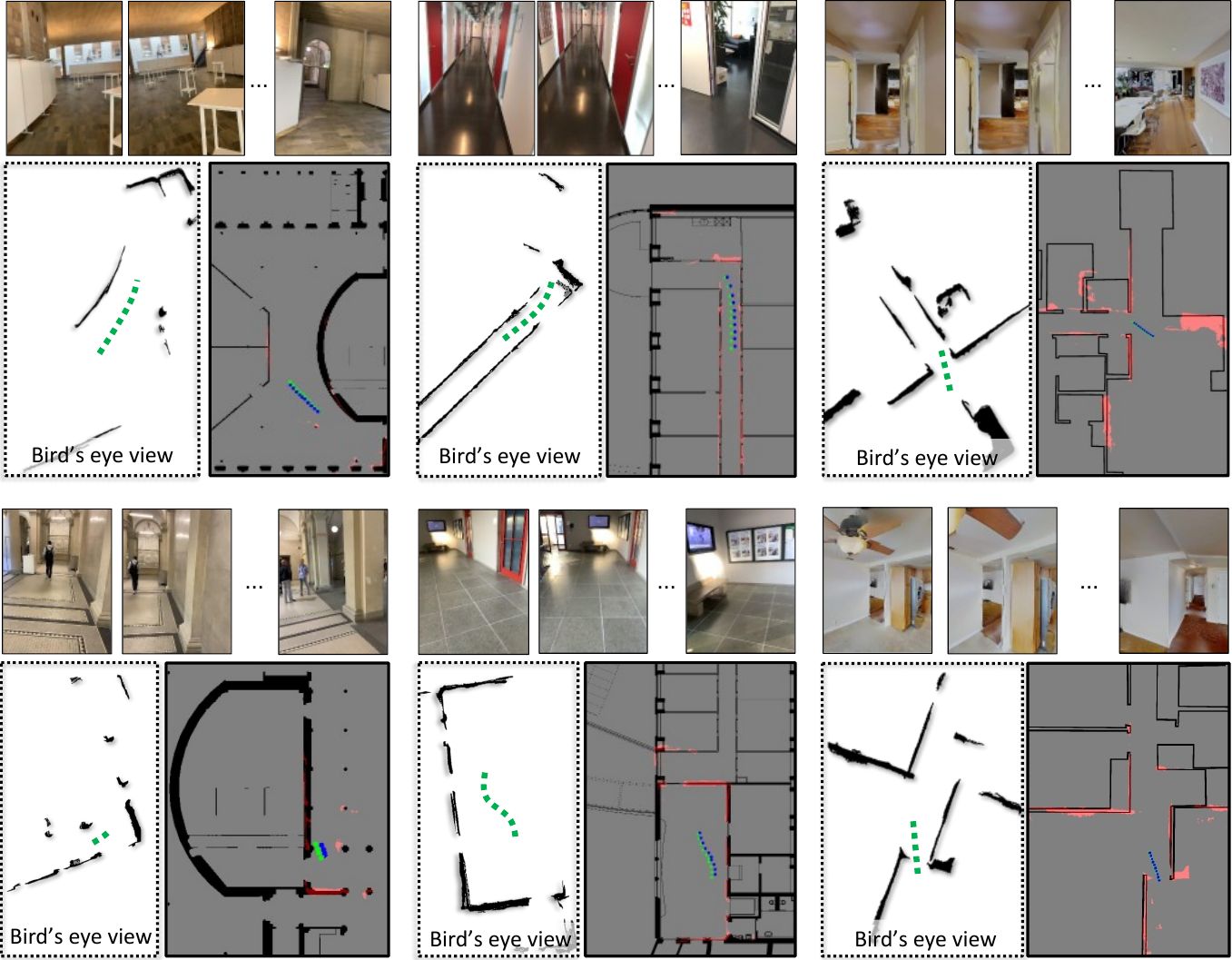}
    \vspace{-2mm}
    \caption{\textbf{Qualitative results.} Each block shows a test scene. \textit{Left}: the input floorplan with the camera trajectory (green). \textit{Right}: the corresponding BEV reconstruction overlaid on the floorplan after alignment with the ground-truth trajectory (blue) and our estimate (green). Our method accurately localizes across diverse environments -- from synthetic rooms (Gibson) to large real-world buildings (LaMAR HGE, CAB) -- without scene-specific training.}
    \label{fig:qualitative}
    \vspace{-5mm}
\end{figure}

\vspace{1mm}
\noindent
\textbf{Baselines.}
We compare with F$^3$Loc~\cite{f3loc} and its successor UnLoc~\cite{unloc}, the current state-of-the-art for floorplan localization.
We use the monocular variant of F$^3$Loc (F$^3$Loc-mono) as the primary baseline and additionally report its fusion variant on Gibson, where monocular depth is combined with multi-view stereo.
For UnLoc, uncertainty estimation is enabled during evaluation.
We also report LASER~\cite{laser} results from~\cite{f3loc} on Gibson for completeness.

\vspace{1mm}
\noindent
\textbf{Metrics.}
We adopt the success rate (SR): a localization run is deemed successful at threshold $X$\,m if the estimated pose remains within $X$\,m of ground truth during the final 10 frames.
SR is reported at varying sequence lengths (\eg, 15, 20, 35, 50, and 100 frames).
We additionally report the root mean square error (RMSE) of trajectory tracking over the last 10 frames, computed over successful runs ($\text{RMSE}_{\text{succ}}$) and over all runs ($\text{RMSE}_{\text{all}}$).

\vspace{1mm}
\noindent
\textbf{Experimental setup.}
For F$^3$Loc, we use the publicly released pretrained model on Gibson(t); since no pretrained LaMAR model is available, we retrain it using the released code.
Among the 16 LaMAR HGE sessions, three are reserved for evaluation and the rest for training.
UnLoc is trained on the respective training splits of Gibson and LaMAR HGE following the same protocol.
For our method, all hyperparameters are fixed across all experiments and determined solely on the LaMAR HGE training split.
The number of RANSAC iterations is set to 500.
We enable all minimal solvers that are geometrically valid per dataset: the 3L and 2L solvers for Gibson and LaMAR CAB (linear wall structures only), and all four solvers for LaMAR HGE (which also contains cylindrical structures).
For our method, we conduct 5 independent runs with different random seeds and report the mean.

    \vspace{-1mm}
\subsection{Main Results}
    \vspace{-1mm}

\textbf{LaMAR HGE} (Tab.~\ref{tab:tab1}).
F$^3$Loc and UnLoc are trained on the HGE training split; our method uses it only for hyperparameter selection.
On the full floorplan, UnLoc achieves perfect SR at 100 frames, while our training-free method reaches 83.6\% -- more than doubling the 36.4\% of trained F$^3$Loc.
The gap between our method and UnLoc narrows at shorter, more challenging sequence lengths (\eg, 44.8\% vs.\ 63.5\% at \#20), where the advantage of sequential filtering in UnLoc becomes less pronounced.
On the cropped floorplan, our method achieves 85.0\% SR at \#100 and surpasses F$^3$Loc at all sequence lengths, notably reaching 73.1\% at \#35 -- close to UnLoc's 88.5\% -- despite requiring no training.
Interestingly, our SR on the cropped floorplan (85.0\%) slightly exceeds that on the full floorplan (83.6\%), suggesting that reducing the search space helps hypothesis selection.
The higher $\text{RMSE}_{\text{succ}}$ of our method (0.49\,m vs.\ 0.25\,m on Full) reflects that, while our geometric alignment is accurate enough to localize reliably, the learned regression in UnLoc achieves finer sub-meter precision when it succeeds.

\vspace{1mm}\noindent
\textbf{LaMAR CAB} (Tab.~\ref{tab:tab2}).
This experiment directly tests cross-environment generalization: all methods use models trained or tuned on HGE and are evaluated on the different CAB building without any adaptation.
F$^3$Loc fails entirely (0\% across all lengths), confirming that its learned features do not transfer across buildings.
UnLoc retains partial performance (50\% at \#100) but degrades sharply at shorter sequences (16.7\% at \#15).
Our method achieves 100\% SR at both \#100 and \#50, and maintains 58.9\% even at the challenging \#15 setting -- a +42.2 percentage point improvement over UnLoc -- demonstrating that geometric primitives provide robust, transferable localization cues.
This result highlights a key practical advantage: deploying our method in a new building \textit{requires only a floorplan}, whereas learned baselines would need to collect training data and retrain, with no guarantee of comparable accuracy.

\vspace{1mm}\noindent
\textbf{Gibson(t)} (Tab.~\ref{tab:tab3}).
On this synthetic benchmark, our method -- without any training -- achieves 99.5\% SR@1m at 100 frames with the lowest $\text{RMSE}_{\text{all}}$ (0.16\,m) and $\text{RMSE}_{\text{succ}}$ (0.11\,m), surpassing all baselines including the trained F$^3$Loc fusion variant (94.6\%, 0.51\,m).
Trained UnLoc achieves the highest SR at shorter sequences (\eg, 81.3\% at \#15 vs.\ our 66.7\%), benefiting from its learned sequential filtering.
To stress-test generalization, we evaluate UnLoc trained on LaMAR HGE (a real-world office environment) on Gibson (synthetic residential scenes).
Even with intrinsic-aligned input to reduce the domain shift, UnLoc reaches only 8.0\% SR at \#100; without this alignment, it drops to 0\%.
Our method achieves 99.5\% on the same data, highlighting the brittle cross-domain behavior of learned approaches versus the robustness of geometry-based matching.
Notably, our $\text{RMSE}_{\text{succ}}$ (0.11\,m) is also the lowest among all methods, indicating that geometric primitive alignment yields not only high recall but also precise localization in environments with well-defined wall structures.
Qualitative results across all datasets are shown in Fig.~\ref{fig:qualitative}.

\begin{table*}[t]
\centering
\scriptsize
\caption{\textbf{Performance comparison on LaMAR HGE.} Success rate at 1\,m (SR@1m) is reported at varying sequence lengths (\#\,=\,number of input frames). F$^3$Loc and UnLoc are trained on the HGE training split; our method uses it only for hyperparameter selection. Best result in \textbf{bold}, second best \underline{underlined}.}
    \vspace{-2mm}
\begin{tabular}{cl|c|ccc|cccc}
\toprule
 & & \multirow{2}{*}[-3pt]{\makecell{Trained\\on HGE}} & \multicolumn{3}{c|}{Overall (\#100)} 
 & \multicolumn{4}{c}{SR@1m (\%) } \\
\cmidrule(lr){4-6} \cmidrule(lr){7-10}
 Scene & Model & 
 & SR@1m $\uparrow$ & $\text{RMSE}_{\text{succ}} \downarrow$ & $\text{RMSE}_{\text{all}} \downarrow$ 
 & \#50 $\uparrow$ & \#35 $\uparrow$ & \#20 $\uparrow$ & \#15 $\uparrow$ \\
\midrule
\addlinespace[-1pt]
\midrule
\multirow{3}{*}{Full} & F$^3$Loc mono \cite{f3loc} & \cmark
& \phantom{9}36.4 & \underline{0.45} & 27.4
& 16.7 & \phantom{7}5.7 & \phantom{5}1.6 & \phantom{5}1.2 \\
& UnLoc~\cite{unloc} & \cmark
& \textbf{100.0} & \textbf{0.25} & \phantom{5}\textbf{0.3}
& \textbf{75.0} & \textbf{74.3} & \textbf{63.5} & \textbf{50.6} \\
& \textbf{Ours} & \xmark
& \phantom{1}\underline{83.6} & 0.49 & \underline{10.8}
& \underline{63.3} & \underline{54.3} & \underline{44.8} & \underline{29.4} \\
\midrule
\addlinespace[-1pt]
\midrule
\multirow{3}{*}{Cropped} & F$^3$Loc mono \cite{f3loc} & \cmark
& \phantom{5}75.0 & 0.53 & 4.93
& 29.4 & \phantom{5}7.7 & \phantom{5}2.2 & \phantom{5}0.0 \\
& UnLoc~\cite{unloc} & \cmark
& \textbf{100.0} & \textbf{0.41} & \textbf{0.41}
& \textbf{94.1} & \textbf{88.5} & \textbf{71.7} & \textbf{62.9} \\
& \textbf{Ours} & \xmark
& \phantom{5}\underline{85.0} & 0.61 & \underline{2.59}
& \underline{68.2} & \underline{73.1} & \underline{43.0} & \underline{38.4} \\

\bottomrule
\end{tabular}

\label{tab:tab1}
\vspace{-0.2cm}
\end{table*}

\begin{table*}[t]
\centering
\scriptsize
\setlength{\tabcolsep}{9pt}
\caption{\textbf{Cross-scene generalization on LaMAR CAB.} All methods use models trained or tuned on LaMAR HGE -- none are retrained on CAB. SR@1m~(\%) at varying sequence lengths (\#\,=\,number of input frames). Best in \textbf{bold}, second best \underline{underlined}.}
    \vspace{-2mm}
\begin{tabular}{l|c|ccccc}
\toprule
& \multirow{2}{*}[-3pt]{\makecell{Trained\\on CAB}} & \multicolumn{5}{c}{SR@1m (\%)}\\
\cmidrule{3-7}
 Model & 
 & \#100 $\uparrow$
 & \#50 $\uparrow$ & \#35 $\uparrow$ & \#20 $\uparrow$ & \#15 $\uparrow$ \\
\midrule

F$^3$Loc mono \cite{f3loc}
&\xmark & \phantom{10}0 & \phantom{10}0 & \phantom{3}0.0 & \phantom{3}0.0 & \phantom{3}0.0 \\

UnLoc \cite{unloc}
&\xmark & \phantom{1}\underline{50} & \phantom{3}\underline{40} & \underline{57.1} & \underline{30.8} & \underline{16.7} \\

\textbf{Ours}
& \xmark & \textbf{100}
& \textbf{100} & \textbf{65.7} & \textbf{64.6} & \textbf{58.9} \\
\bottomrule
\end{tabular}

\label{tab:tab2}
\vspace{-6mm}
\end{table*}

\begin{table*}[t]
\centering
\scriptsize
\caption{\textbf{Performance comparison on Gibson(t).} Top block: methods trained on Gibson. Bottom block: zero-shot methods (no Gibson training). \#\,=\,number of input frames. Best in \textbf{bold}, second best \underline{underlined} for each category.}
    \vspace{-2mm}
\begin{tabular}{l|c|ccc|cccc}
\toprule
 & & \multicolumn{3}{c|}{Overall (\#100)} 
 & \multicolumn{4}{c}{SR@1m (\%)} \\
\cmidrule(lr){3-5} \cmidrule(lr){6-9}
 Model & Trained
 & SR@1m$\uparrow$ & $\text{RMSE}_{\text{succ}}\downarrow$ & $\text{RMSE}_{\text{all}}\downarrow$ &
 \#50$\uparrow$ & \#35$\uparrow$ & \#20$\uparrow$ & \#15$\uparrow$ \\
\midrule
\addlinespace[-1pt]
\midrule

LASER \cite{laser}
& \cmark & 59.5 & 0.39 & 1.96 
& -- & -- & -- & -- \\
F$^3$Loc mono \cite{f3loc}
& \cmark & 89.2 & 0.18 & 0.88 
& 70.5 & 55.9 & 34.0 & 28.4 \\

F$^3$Loc fusion \cite{f3loc}
& \cmark & \underline{94.6} & \textbf{0.12} & \underline{0.51} 
& \underline{84.6} & \underline{69.4} & \underline{46.0} & \underline{41.8} \\

UnLoc \cite{unloc}
& \cmark & \textbf{97.3} & \underline{0.16} & \textbf{0.28} 
& \textbf{94.9} & \textbf{92.8} & \textbf{86.5} & \textbf{81.3} \\

\midrule
UnLoc (original input) \cite{unloc}
& \xmark & \phantom{1}0.0 & -- & 5.54 & \phantom{1}1.0
 & \phantom{1}1.0 & \phantom{1}0.0 & \phantom{1}0.0 \\
UnLoc (intr-aligned input) \cite{unloc}
& \xmark & \phantom{1}\underline{8.0} & \underline{0.60} & \underline{4.83} & \phantom{1}\underline{9.0}
 & \phantom{1}\underline{5.0} & \phantom{1}\underline{1.0} & \phantom{1}\underline{1.0} \\
\textbf{Ours} 
& \xmark & \textbf{99.5} & \textbf{0.11}& \textbf{0.16}
 & \textbf{91.0} & \textbf{86.5} & \textbf{71.4} & \textbf{66.7} \\
\bottomrule
\end{tabular}

\label{tab:tab3}
\vspace{-4mm}
\end{table*}

    \vspace{-1mm}
\subsection{Ablation Study}
\label{subsec:ablation}
    \vspace{-1mm}

We ablate the major pipeline components on LaMAR HGE (full) in Tab.~\ref{tab:tab4}, reporting the mean over five random seeds.

\vspace{1mm}\noindent
\textbf{Reconstruction backbone.}
Replacing MapAnything with Depth Anything 3~\cite{depthanything3} confirms that our framework generalizes across backbones.
The giant variant even surpasses MapAnything at longer sequences (90.9\% vs.\ 84.9\% at \#100; 69.4\% vs.\ 59.7\% at \#50), likely due to more accurate metric depth enabling better BEV reconstructions when sufficient views are available.
At mid-range lengths, MapAnything's direct BEV prediction is more stable (45.0\% vs.\ 35.5\% at \#20), suggesting that depth-based reconstruction is more sensitive to limited viewpoint coverage.
The base variant trades accuracy for efficiency (60.6\% at \#100), demonstrating a graceful performance-capacity trade-off.

\vspace{1mm}\noindent
\textbf{Solver composition.}
The four solvers provide complementary cues.
The 2C solver alone reaches 72.7\% at \#100 but drops sharply to 12.9\% at \#15, since short sequences rarely capture enough arc coverage for reliable circle fitting.
Conversely, line-based solvers (2L+3L) achieve 20.8\% at \#15 -- nearly twice as high -- because lines are ubiquitous in man-made environments and detectable even in few frames.
Adding the LC solver to the line-based set (2L+3L+LC) boosts \#15 from 20.8\% to 31.4\%, as the mixed line-circle constraint offers additional diversity.
Combining all four solvers yields the best overall performance, validating the hybrid RANSAC strategy.

\vspace{1mm}\noindent
\textbf{Feature-based matching.}
We replace our geometric primitives with correspondences from SIFT~\cite{sift}, SuperPoint+LightGlue~\cite{superpoint, lightglue}, and RoMa v2~\cite{romav2}, integrated into the same estimation framework.
All feature-based methods fail under the severe cross-modal domain gap between BEV and floorplans, confirming that geometric primitives are essential for training-free localization.

\begin{table}[t]
\centering
\scriptsize
\setlength{\tabcolsep}{5pt}
\caption{\textbf{Ablation study on LaMAR HGE (full).} Each row disables or replaces one component while keeping the rest fixed. SR@1m (\%) is reported at varying sequence lengths. The top row shows the full pipeline.}
    \vspace{-2mm}
\begin{tabular}{ll|ccccc}
\toprule
& & \multicolumn{5}{c}{SR@1m (\%)} \\
\cmidrule(lr){3-7}
  Component & Configuration & \#100 $\uparrow$ & \#50 $\uparrow$ & \#35 $\uparrow$ & \#20 $\uparrow$ & \#15 $\uparrow$ \\
\midrule
All & (default) & 84.9 & 59.7 & 55.2 & 45.0 & 30.2 \\
\midrule
\addlinespace[-1pt]
\midrule
\multirow{3}{*}{\makecell[l]{Base\\model}} & MapAnything (default) & 84.9 & 59.7 & 55.2 & 45.0 & 30.2 \\
 & DepthAnything 3 giant \cite{depthanything3} & 90.9 & 69.4 & 53.3 & 35.5 & 36.1 \\
 & DepthAnything 3 base \cite{depthanything3} & 60.6 & 62.5 & 41.9 & 29.6 & 19.6 \\
\midrule
\multirow{3}{*}{Solvers} & Only 2C & 72.7 & 45.8 & 31.4 & 17.5 & 12.9 \\
 & 2L+3L+LC & 66.7 & 56.9 & 50.5 & 38.1 & 31.4 \\
 & 2L+3L & 57.6 & 56.9 & 41.0 & 27.0 & 20.8 \\
\midrule
\multirow{3}{*}{Features} & SIFT \cite{sift} & 0.0 & 4.2 & 0.0 & 0.0 & 0.0 \\
 & SP+LG \cite{superpoint, lightglue} & 0.0 & 0.0 & 2.9 & 0.0 & 0.0 \\
 & RoMa v2 \cite{romav2} & 9.1 & 12.5 & 17.1 & 4.8 & 3.5 \\
\midrule
\multirow{2}{*}{\makecell[l]{Score\\function}} & Only $C^c$ & 60.6 & 45.8 & 18.1 & 5.8 & 5.9 \\
 & Only $C^v$ & 6.1 & 25.0 & 33.3 & 29.6 & 23.9 \\
\midrule
Refinement & Disabled & 63.6 & 58.3 & 55.2 & 45.5 & 25.9 \\
\bottomrule
\end{tabular}
\label{tab:tab4}
    \vspace{-4mm}
\end{table}

\vspace{1mm}\noindent
\textbf{Score function.}
The two scoring components exhibit strongly asymmetric failure modes.
$C^c$ alone achieves 60.6\% at \#100 but collapses to 5.9\% at \#15: in dense wall regions, many incorrect hypotheses produce high consistency scores, and with fewer frames the ranking signal becomes unreliable.
$C^v$ alone shows the opposite pattern -- only 6.1\% at \#100 but 23.9\% at \#15 -- because it penalizes violation but does not reward positive alignment, so longer sequences dilute its discriminative power.
Combining both scores captures complementary cues: $C^c$ rewards correct alignment while $C^v$ penalizes geometric violation, together yielding robust hypothesis selection across all sequence lengths.

\vspace{1mm}\noindent
\textbf{Refinement.}
Minimal-solver hypotheses are inherently noisy.
Since Success Rate requires all of the last 10 frames to fall within 1\,m, even small inaccuracies cause failure.
Levenberg--Marquardt refinement has the largest impact at longer sequences (63.6\%$\to$84.9\% at \#100), where more wall points provide a richer objective for optimization.
At shorter sequences, the effect diminishes (25.9\%$\to$30.2\% at \#15) because fewer observations limit the refinement landscape, yet the gain remains consistent.

\begin{table*}[t]
\centering
\scriptsize
\caption{\textbf{Runtime analysis.} SR@1m (\%) and per-component runtime (seconds) as a function of the number of RANSAC iterations on LaMAR HGE (full, all four solvers) and Gibson(t) (2L+3L solvers). Accuracy saturates early while runtime grows linearly with the iteration count.}
    \vspace{-2mm}
\setlength{\tabcolsep}{3pt}
\centering
\begin{tabular}{@{}c|cccc|cc@{}}
\toprule
& \multicolumn{4}{c}{LaMAR HGE full (3L+2L+2C+LC)} & \multicolumn{2}{c}{Gibson(t) (2L+3L)} \\
\cmidrule(lr){2-5}\cmidrule(lr){6-7}
 \# Iteration & SR@1m (\%) $\uparrow$ & \makecell[l]{Runtime (s) \\ @ model est. $\downarrow$} & 
    \makecell[l]{Runtime (s) \\ @ scoring $\downarrow$} & Total (s) $\downarrow$ & SR@1m (\%) $\uparrow$ & Total (s) $\downarrow$ \\
\midrule
100  & 87.9 & 0.28 & 2.95 & \phantom{1}4.8 & 99.1 & 0.41 \\
500 & 84.9 & 1.4 & 3.3 & 6.3 & 99.1 & 0.51 \\
1000 & 84.9 & 2.51 & 4.08 & 7.9 & 99.1 & 0.60\\
5000 & 84.9 & 13.6 & 12.2 & 26.6 & 96.4 & 0.95\\
\bottomrule
\end{tabular}

\vspace{4pt}
\label{tab:tab6}
\vspace{-0.5cm}
\end{table*}

\vspace{1mm}
\noindent
\textbf{Runtime.}
Tab.~\ref{tab:tab6} shows SR and per-component runtime as a function of the RANSAC iteration count.
Accuracy saturates early (\eg, 100--300 iterations), while runtime grows linearly.
Excessive iterations produce redundant hypotheses with marginal score differences, yielding diminishing returns.
We set the default to 500 iterations as a conservative trade-off.

\begin{figure}[t]
    \centering

    \begin{subfigure}{0.49\linewidth}
        \centering
        \includegraphics[width=\linewidth]{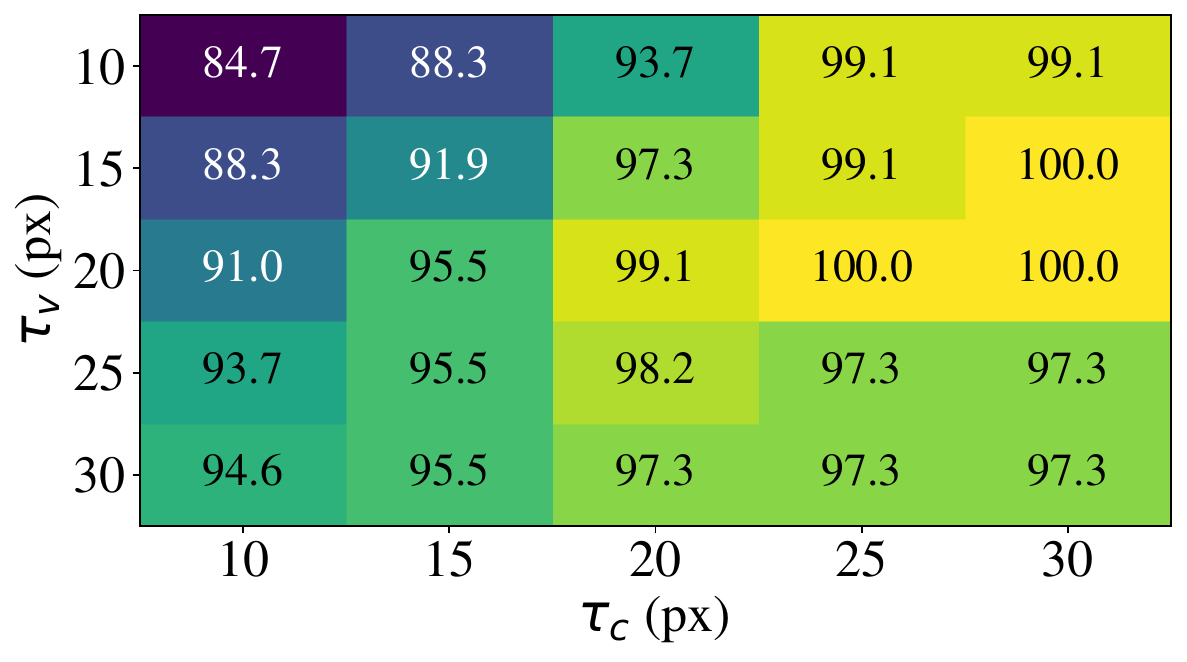}
        \caption{Gibson(t)}
    \end{subfigure}
    \hfill
    \begin{subfigure}{0.49\linewidth}
        \centering
        \includegraphics[width=\linewidth]{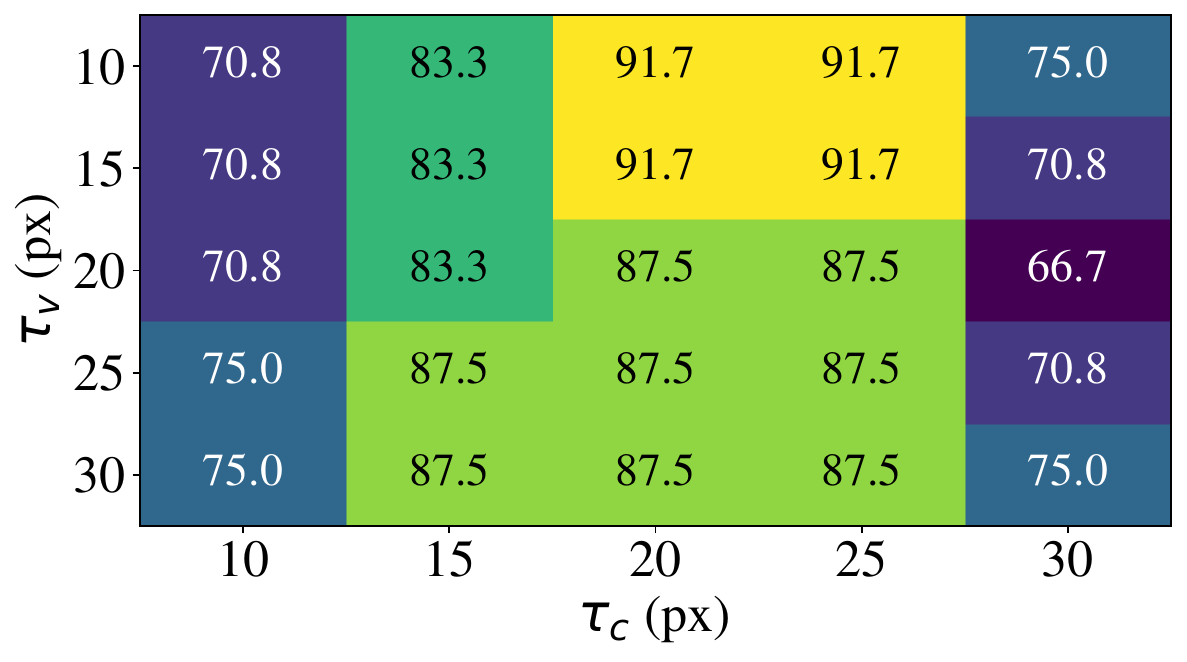}
        \caption{LaMAR HGE cropped}
    \end{subfigure}

    \caption{\textbf{Parameter sensitivity.} Each cell shows SR@1m~(\%) at sequence length 100 for a given combination of consistency threshold $\tau_c$ (columns) and violation threshold $\tau_v$ (rows). (a)~On Gibson(t), performance is stable across all settings. (b)~On LaMAR HGE (cropped), sensitivity is higher due to the structural complexity of the floorplan, yet strong accuracy is maintained across a wide operating range.}
    \label{fig:sensitivity}
\end{figure}

\vspace{1mm}
\noindent
\textbf{Parameter sensitivity.}
Fig.~\ref{fig:sensitivity} varies the scoring thresholds $\tau_c$ and $\tau_v$ on Gibson(t) and LaMAR HGE (cropped).
On Gibson, performance is highly stable across all settings, as the refinement stage can compensate for coarse initial alignment in sparse environments.
On the more complex LaMAR HGE floorplans, sensitivity is higher -- tight thresholds prune valid hypotheses, while loose thresholds admit false positives -- yet strong performance is maintained across a wide operating range.

\section{Conclusion}
    
We presented Z-FLoc, a zero-shot floorplan localization pipeline that requires no environment-specific training.
Our approach extracts geometric primitives -- lines and circles -- from monocular BEV reconstructions and matches them to floorplan geometry via a family of dedicated minimal solvers within a hybrid RANSAC framework, coupled with a complementary scoring function that combines alignment consistency with free-space violation rejection.
Experiments on Gibson(t) and LaMAR demonstrate that our method achieves state-of-the-art results in cross-environment generalization, reaching 100\% success rate on unseen buildings where trained baselines degrade significantly -- all with a single, fixed set of hyperparameters.
These results suggest that structural geometric reasoning provides a robust and practical alternative to learned representations for floorplan-based localization.

\vspace{1mm}
\noindent
\textbf{Limitations.}
Runtime is dominated by dense 3D reconstruction; incremental processing and GPU acceleration (all experiments use CPU only) could substantially reduce it.
Symmetric structures (\eg, corridors) may yield multiple high-scoring hypotheses -- a multi-modal filtering extension would address this.
Finally, the fixed scoring thresholds $\tau_c$ and $\tau_v$ may benefit from adaptive selection in structurally diverse environments.


%
%
\bibliographystyle{splncs04}
\bibliography{main}
\end{document}